\documentclass[12pt,letterpaper]{article}
\usepackage[a4paper,total={7in,10in}]{geometry}

\usepackage{graphicx}
\usepackage{helvet}
\usepackage{authblk}
\usepackage{amsmath}
\usepackage{amssymb}
\usepackage{booktabs}
\usepackage{multirow}
\usepackage{url}
\usepackage{hyperref}
\usepackage[super,comma,sort&compress]{natbib}

\makeatletter
\renewcommand{\maketitle}{\bgroup\setlength{\parindent}{0pt}
\begin{flushleft}
  \textbf{\@title}

  \@author
\end{flushleft}\egroup}
\makeatother

\title{Participatory provenance as representational auditing for AI-mediated public consultation}
\date{}
\author[1,*]{Sachit Mahajan}
\affil[1]{Computational Social Science, Department of Humanities, Social and Political Sciences, ETH Zurich, R\"amistrasse 101, 8092 Zurich, Switzerland}
\affil[*]{Correspondence: sachit.mahajan@gess.ethz.ch}

\begin{document}
\maketitle

\section*{SUMMARY}

AI-assisted consultation can speed large-scale public engagement, but concise summaries may reflect some submissions more closely than others. This paper introduces participatory provenance, a framework for auditing how semantic coverage is distributed from submissions to summary sentences. Applied to two topics in Canada’s 2025 AI Strategy consultation (5,253 records; 2,861 participants), official summaries had higher observed mean coverage than exact-length random text, although statistical significance depended on the embedding model. Low coverage concentrated in semantic regions, especially those centered on criticism of educational technology and distrust of technology and oversight, whereas few or no records crossed the operational threshold in several better-covered regions. Same-budget, cross-fitted extractive benchmarks improved mean and lower-tail coverage on held-out submissions, showing that better semantic coverage was feasible without longer summaries. Consultation summaries should be evaluated not only for coherence and factual support, but also for how coverage is distributed across the range of submitted views.

\section*{KEYWORDS}

participatory AI; algorithmic accountability; public consultation; democratic representation

\section*{INTRODUCTION}
\label{sec:intro}

Whose views appear in an official summary of public input can shape what a government treats as a priority. Democratic and deliberative traditions regard voice, inclusion, and the design of participation as central to legitimate governance~\citep{arnstein1969ladder,pateman1970participation,habermas1984theory,dryzek2000deliberative,fishkin2009speaking,nabatchi2012democracy,gastil2014democracy,fung2006varieties}. Public consultation was once limited by how much material human analysts could read and synthesize. Natural-language processing now allows institutions to process thousands of submissions more quickly. This change creates a basic accountability question: does a concise summary preserve only the most common themes, or does it also reflect the range of views that people submitted?

Most existing checks focus on the output. Explainable-AI methods examine why a system produced an answer~\citep{lundberg2017unified,ribeiro2016should,kim2018interpretability, mahajan2025mapping}. Grounding and hallucination methods ask whether generated claims are supported by a source~\citep{lewis2020retrieval,gao2023retrieval,ji2023survey,manakul2023selfcheck}. Model cards, datasheets, and data-provenance systems document sources, transformations, and intended uses~\citep{mitchell2019model,gebru2021datasheets,buneman2001provenance,simmhan2005survey,herschel2017survey}. These checks are important as foundation models make automated text production easier and more fluent~\citep{vaswani2017attention,bender2021stochastic,bommasani2021opportunities}. They do not, however, show whether a summary is much closer to some submissions than to others. A summary can be supported by the source material and still represent the inputs unevenly.

That gap matters in participatory governance. Summarization systems are commonly designed to produce coherent text that covers frequent content, not to distribute representation evenly across all submissions~\citep{nallapati2017summarunner,see2017get}. Less common views, including views that challenge the premise of a consultation, may therefore be difficult to find in the final text. Research on participatory AI has shown why power and meaningful involvement matter, and has documented disparities in classification, search, and other settings~\citep{sloane2020participation,birhane2022power,delgado2023participatory,young2000inclusion,buolamwini2018gender,chouldechova2017fair,corbettdavies2017algorithmic,raghavan2020mitigating,bartlett2022consumer,noble2018algorithms,oneil2016weapons,selbst2019fairness,raji2020closing}. Work on accountable metrics further supports the need for measurements that match the setting in which they are used~\citep{xia2024towards,mahajan2025mapping,mahajan2026revisiting}. When a short summary gives consistently lower coverage to some parts of the source material, its apparent consensus may conceal uneven representation. Measuring that pattern is necessary before its causes or consequences can be assessed.

\begin{figure}[ht]
  \centering
  \includegraphics[width=\linewidth]{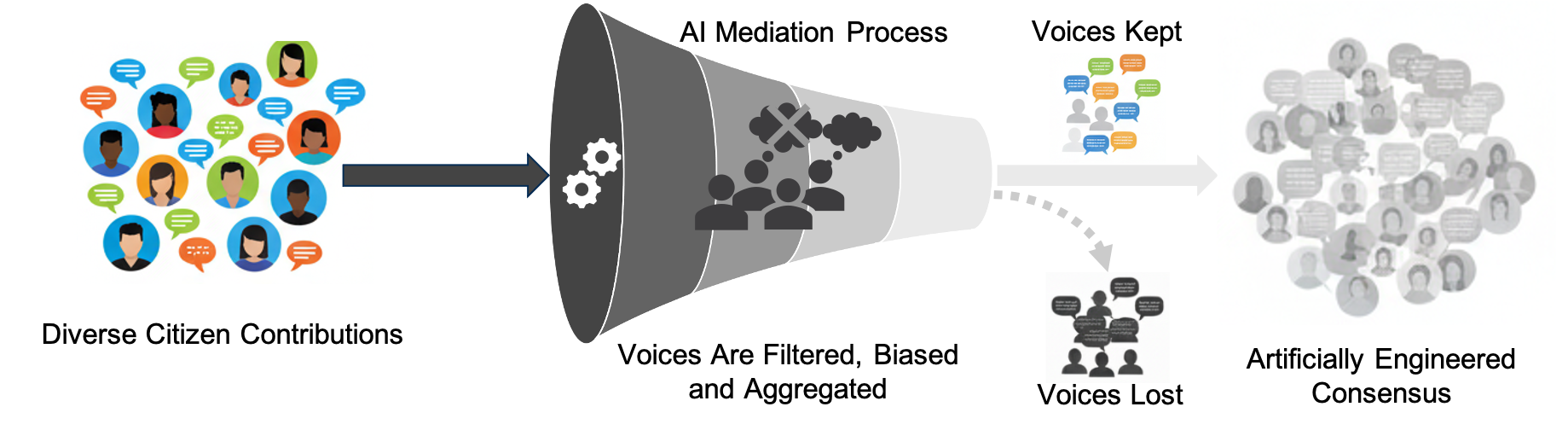}
  \vspace{-0.4cm}
  \caption{\textbf{The apparent-consensus problem in AI-mediated public consultation.} Diverse submissions enter a compression process. A coherent and factually grounded output can retain common views while giving consistently low semantic coverage to other regions of the input distribution. Participatory provenance audits the input-to-summary transformation and reports both aggregate and lower-tail performance.}
  \label{fig:consensus}
\end{figure}

This work introduces \textbf{participatory provenance}, a framework for auditing the path from many submissions to one summary (Fig.~\ref{fig:consensus}). The framework gives each participant a semantic coverage score based on the summary sentence closest to that participant's submission. Optimal Transport, implemented as Wasserstein-2 distance, measures the overall geometric difference between the distribution of participant submissions and the distribution of summary sentences~\citep{kantorovich1942translocation,arjovsky2017wasserstein,courty2017joint,schiebinger2019optimal}. This distance can be understood as the minimum geometric ``effort'' needed to transport one embedding distribution to the other; it is not a measure of editorial or cognitive effort. Coverage distributions and cluster-level rates show where low coverage is concentrated. An exact-length random benchmark asks whether the official summary performs better than chance. A cross-fitted extractive benchmark asks how much coverage is feasible with the same six sentence-level word limits. Evaluation on held-out participants tests whether the benchmark advantage persists beyond the records used to select its sentences.

The framework is applied to Canada's 2025 consultation on its next AI Strategy and the corresponding ISED summary report~\citep{canada2025aistrategy}. The consultation forms part of a wider move toward formal citizen participation and risk-based AI governance, reflected in OECD guidance, Canada's proposed AIDA, and the EU AI Act~\citep{oecd2020innovative,canada2022aida,eu2024aiact}. These instruments address participation, data governance, risk management, or transparency, but none specifies a participant-level audit of how semantic coverage is distributed across source submissions. The analysis covers Education \& Skills ($n=2{,}496$) and Safe AI \& Public Trust ($n=2{,}757$), including 2{,}392 respondents who contributed to both topics, each topic paired with its six-sentence official summary. Because the available data do not identify which model or human decision produced a given sentence, the audit addresses the published relationship between submissions and summaries rather than ISED's internal drafting process.

The audit shows that low coverage was concentrated rather than evenly spread. It accumulated in semantic regions centered on criticism of educational technology and on distrust of technology and oversight, although it did not follow a simple divide between supportive and critical views, and respondents poorly covered in one topic tended to be poorly covered in the other (odds ratio 6.54). The official summaries had higher observed mean coverage than exactly length-matched random text, a difference that was statistically significant under only one of the two embedding models tested, and no lower-tail comparison reached significance. Held-out extractive benchmarks under the same sentence-level word limits found that better mean and lower-tail coverage was attainable. Average fidelity is therefore an incomplete description of a summary; the distribution of coverage across the source material should be reported alongside it. 

\section*{RESULTS}
\label{sec:results}

\subsection*{Average coverage conceals a poorly covered minority}

Responses to up to three sub-questions were combined within respondent and topic. Preprocessing removed French-language records, texts shorter than five words, spam or placeholders, and a small number of off-topic records (Methods; Table S1). The final corpus contains 2{,}496 Education records and 2{,}757 Trust records. Participant texts and the six official sentences for each topic were embedded with text-embedding-3-large.

\begin{figure}[ht!]
  \centering
  \includegraphics[width=\linewidth]{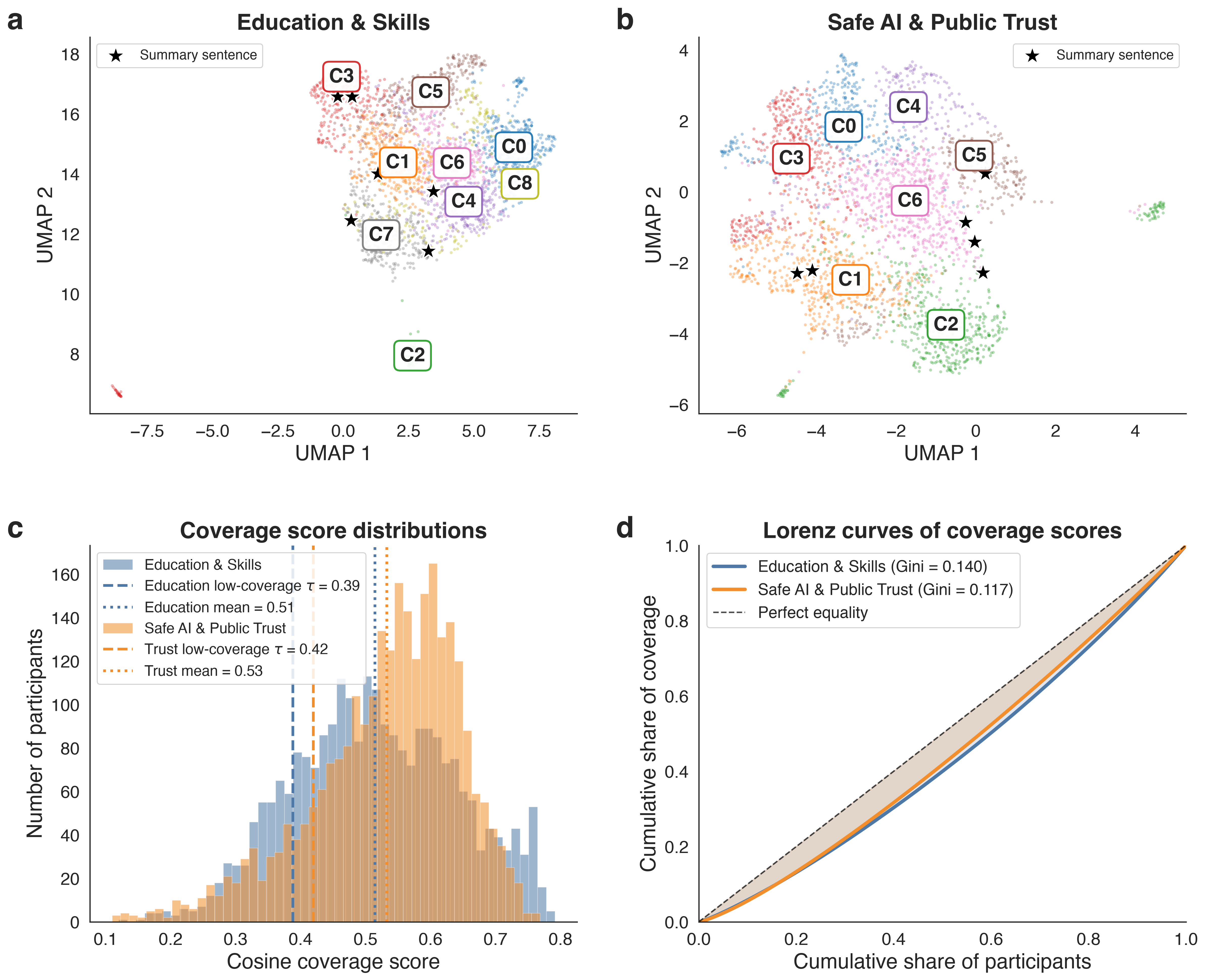}
  \vspace{-0.4cm}
  \caption{\textbf{Participatory provenance overview.} \textbf{a,b,} UMAP projections of respondent--topic embeddings, colored by the selected analytic partition; labels C0--C8 in Education and C0--C6 in Trust are cluster identifiers, and stars indicate official-summary sentences. UMAP is used only for visualization; the cluster assignments, coverage scores, and statistical analyses were not calculated in this two-dimensional space. \textbf{c,} Coverage-score distributions. Dashed lines show the operational low-coverage thresholds ($\tau=\bar c-\sigma_c$: 0.388 Education; 0.419 Trust), and dotted lines show means (0.514 and 0.533). \textbf{d,} Lorenz curves computed on raw non-negative coverage scores. The diagonal represents equal coverage scores; greater departure from it and larger Gini values indicate greater coverage inequality. Gini coefficients are 0.140 (95\% bootstrap CI [0.136, 0.144]) and 0.117 [0.113, 0.121].}
  \label{fig:overview}
\end{figure}

For each respondent--topic record, semantic coverage was the largest cosine similarity between its embedding and the embeddings of the six official-summary sentences, as formally defined in Methods. In plain terms, each record received the similarity score of its nearest official-summary sentence; larger values indicate a closer semantic match. Mean coverage was 0.514 (95\% bootstrap CI [0.509, 0.519]) for Education and 0.533 [0.528, 0.537] for Trust. The corresponding Wasserstein-2 distances in PCA-50 space were 0.705 and 0.684. Raw-score coverage Ginis were 0.140 and 0.117 (Fig.~\ref{fig:overview}c,d). These values describe the overall distributions, but they do not show where low coverage occurs.

The operational low-coverage rule, $c(i)<\tau$ with $\tau=\bar c-\sigma_c$, identified 421 Education records (16.9\%, 95\% CI [15.3\%, 18.4\%]) and 423 Trust records (15.3\%, [14.0\%, 16.7\%]). This threshold is relative to each topic's coverage distribution, so the overall percentage depends partly on the rule. The more informative result is how strongly low coverage was concentrated in particular parts of each corpus.

To locate those parts, the Education corpus was represented by nine $k$-means regions and Trust by seven (Fig.~\ref{fig:overview}a,b; Figure S1). The selected partitions were reproducible under resampling (mean adjusted Rand index 0.843 and 0.869), but their separation was weak: the best silhouette scores were 0.079 and 0.076, and mean normalized pointwise mutual information (NPMI) was $-0.807$ and $-0.619$. The records therefore form continuous semantic fields rather than clear-cut groups. Cluster labels briefly describe the documents nearest each centroid. Cluster results are used to locate patterns in the corpus, not to define discrete social groups.

\subsection*{Better coverage was feasible within the same word budget}
\label{subsec:coverage}

\begin{figure}[ht!]
  \centering
  \includegraphics[width=\linewidth]{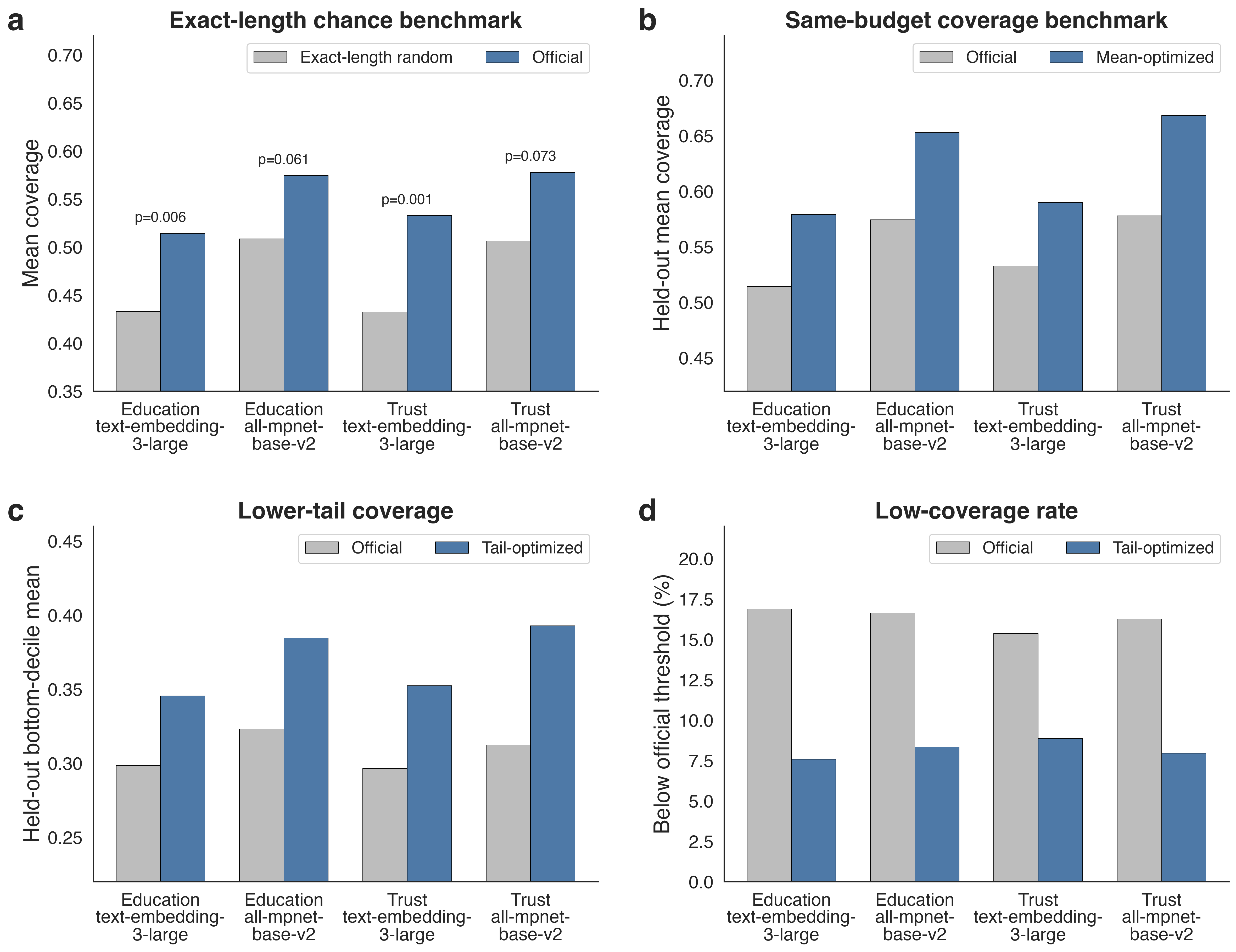}
  \caption{\textbf{Chance and same-budget benchmarks answer different questions.} \textbf{a,} Official mean coverage versus 1{,}000 exact-length random pseudo-summaries; displayed $p$-values are unadjusted one-sided empirical probabilities. \textbf{b,} Official versus mean-optimized extractive coverage on held-out participants. \textbf{c,} Official versus tail-optimized mean coverage among the 10\% lowest-covered held-out records. \textbf{d,} Proportion below the official summary's fixed threshold $\tau$. Cross-fitted values average participant-level out-of-fold predictions across five repetitions. The models are text-embedding-3-large and all-mpnet-base-v2.}
  \label{fig:benchmarks}
\end{figure}

The first benchmark asked whether the official summaries covered the source material better than random text of exactly the same length. Each of 1{,}000 pseudo-summaries contained six contiguous participant-text windows with word counts matched to the six official sentences. The observed official mean was higher in all four topic and model combinations (Fig.~\ref{fig:benchmarks}a). With text-embedding-3-large, Education coverage was 0.514 for the official summary and 0.433 for the random benchmark ($p=0.006$); Trust coverage was 0.533 and 0.432 ($p=0.001$). With all-mpnet-base-v2, the observed differences were similar in direction but neither met a 0.05 threshold (0.574 versus 0.508, $p=0.061$; 0.578 versus 0.506, $p=0.073$). The official bottom-decile mean was also higher in all four observed comparisons, but none was individually significant ($p=0.107$--0.366; Table S6). The random benchmark is therefore a chance floor, not evidence of overall summary quality.

The second benchmark asked whether better coverage was possible within the same six sentence-level word limits. Candidate summaries used six complete participant sentences, with each sentence no longer than the corresponding official sentence. Duplicate candidates and exact repetitions of consultation prompts were excluded, and every selected sentence came from a different participant. Selection used only training participants, followed by evaluation on held-out participants in five repetitions of shuffled five-fold cross-fitting. One objective selected sentences to improve mean coverage. The other balanced mean coverage with bottom-decile coverage. These extracts were designed only as coverage benchmarks; coherence, redundancy, policy usefulness, and writing quality were not optimized.

The same-budget benchmarks consistently exceeded the official summaries on held-out records (Fig.~\ref{fig:benchmarks}b--d; Table S7). Mean-optimized coverage increased by 0.065 (95\% conditional participant-bootstrap CI [0.063, 0.067]) for Education with text-embedding-3-large, 0.078 [0.076, 0.081] for Education with all-mpnet-base-v2, 0.057 [0.055, 0.059] for Trust with text-embedding-3-large, and 0.090 [0.088, 0.093] for Trust with all-mpnet-base-v2. Tail-oriented selection improved the bottom-decile mean by 0.047 [0.041, 0.053], 0.061 [0.054, 0.069], 0.056 [0.049, 0.063], and 0.081 [0.072, 0.089], respectively. All eight conditional intervals excluded zero. One-sided paired-randomization $p$-values were 0.0002 before and 0.0016 after Holm adjustment. Every comparison improved in all 25 folds, and all gains were positive in each of the five repeated partitions (Figure S2; Table S8). Better coverage was therefore possible without increasing the word budget, although the benchmark does not establish that its selected sentences would form a suitable public summary.

\subsection*{Low coverage concentrates in specific semantic regions}
\label{subsec:clusters}

The terms \textit{low coverage} and \textit{operationally excluded} refer only to the defined criterion $c(i)<\tau$; they do not imply intentional removal or describe a participant's experience. The question is where low coverage concentrated and whether that pattern persisted under alternative thresholds (Fig.~\ref{fig:robustness}d).

\begin{figure}[ht!]
  \centering
  \includegraphics[width=\linewidth]{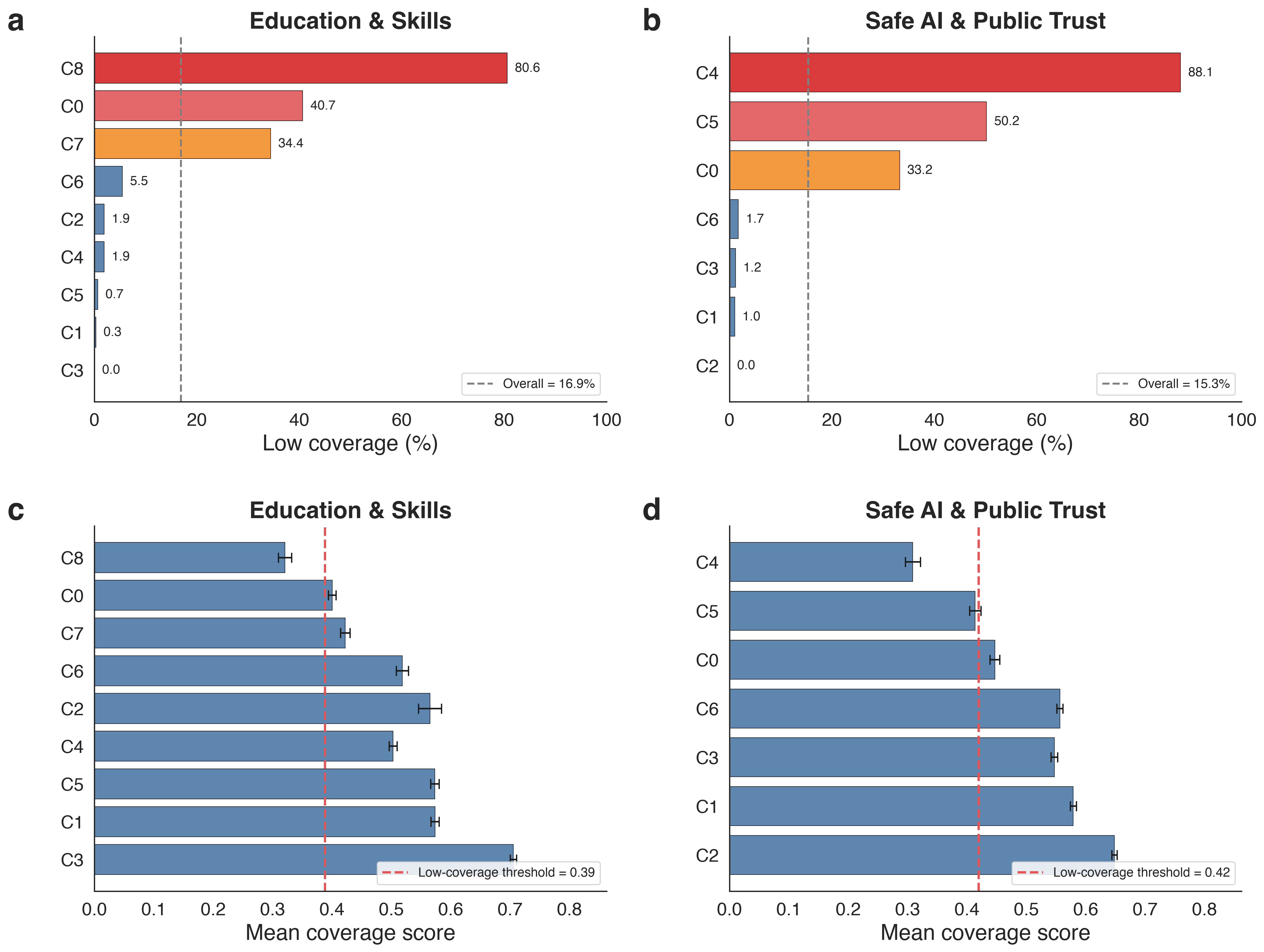}
  \vspace{-0.3cm}
  \caption{\textbf{Low coverage is concentrated in selected semantic regions.} \textbf{a,b,} Operational low-coverage rate by cluster, sorted within topic; the overall rates are 16.9\% for Education and 15.3\% for Trust. \textbf{c,d,} Cluster mean coverage; dashed lines show the topic-specific thresholds $\tau$. Error bars are 95\% normal intervals for the cluster means. Labels are abbreviated in the figure, with full labels in Tables S2--S3. They describe centroid-nearest documents and do not define social groups.}
  \label{fig:clusters}
\end{figure}

In Education, C8 (\textit{Critique of Educational Technology Implementation}; $n=165$) had mean coverage of 0.321, and 80.6\% of its records met the low-coverage criterion (Fig.~\ref{fig:clusters}). The rates were also high in C0 (\textit{Opposition to AI in Society}; $n=386$) at 40.7\% and C7 (\textit{Holistic Education and Community Engagement}; $n=311$) at 34.4\%. In contrast, the rates were 0.0\% in C3 (\textit{Workforce Skills for Digital Economy}; $n=336$), 0.3\% in C1 (\textit{AI Integration in Education Systems}; $n=369$), and 0.7\% in C5 (\textit{Scepticism towards AI in education}; $n=302$). The absolute difference between C8 and C3 was 80.6 percentage points.

In Trust, C4 (\textit{Distrust in technology and oversight}; $n=193$) had mean coverage of 0.309 and a low-coverage rate of 88.1\%. The rate was 50.2\% in C5 (\textit{Trust through Transparency and Regulation}; $n=251$) and 33.2\% in C0 (\textit{Rejection of AI and its implications}; $n=322$). It was 0.0\% in C2 (\textit{Building public trust in AI}; $n=501$). The gap between the best-covered and worst-covered regions was accordingly large: 80.6 percentage points in Education (C3 at 0.0\% versus C8 at 80.6\%) and 88.1 percentage points in Trust (C2 at 0.0\% versus C4 at 88.1\%). Criticism alone did not determine coverage: one skeptical Education region had a low rate, whereas the Trust region centered on distrust had the largest shortfall.

Tail-oriented selection reduced, but did not remove, the largest cluster shortfalls. With text-embedding-3-large, the low-coverage rate in Education C8 fell from 80.6\% to 66.7\%, and the rate in Trust C4 fell from 88.1\% to 63.7\%. With all-mpnet-base-v2, the corresponding changes were 78.2\% to 58.2\% and 89.6\% to 50.8\% (Table S9). These post hoc regional diagnostics show that the lower tail partly reflects which content receives space in the summary. They also show that six sentences were not enough to remove the largest differences under the tested objective.

\subsection*{Coverage flows through one or two summary sentences}
\label{subsec:sentenceload}

Assigning each record to its highest-similarity official sentence shows how unevenly the six sentences share the representational load (Table S14). In Education, a single sentence (S5, on a national AI literacy strategy and public campaigns) was the nearest sentence for 66.9\% of all records, whereas S3 and S4 together were nearest for 3.5\%, at mean assigned similarities of 0.39--0.44. In Trust, the load was more balanced: S1 (transparency and governance) and S6 (skepticism toward generative AI) were nearest for 36.3\% and 30.5\% of records. Trust also had higher mean coverage and a lower coverage Gini, a pattern consistent with more evenly distributed sentence load, although two topics cannot establish that relationship.

The assignment also gives a sentence-level view of the worst-covered Trust region. S6, which acknowledges skepticism toward generative AI, was the nearest sentence for 55 of C4's 193 records (28.5\%; S1 for 54), yet C4's mean similarity to S6 was 0.274 across all its records and 0.303 even among the records assigned to it, against a topic-wide mean coverage of 0.533. Within this measure, being a region's most frequent nearest sentence did not correspond to high proximity: thematic acknowledgment and measured coverage are different quantities.

The benchmark selections describe where the improved summaries drew their sentences. The tail-oriented objective selected no sentences owned by Education C8 records in either model's 150 selection instances, yet C8's low-coverage rate fell (Table S14). Because each selected sentence is attributed to the cluster of its owner's full record, and no ablation isolates which sentences produced the C8 improvement, this sourcing pattern is descriptive. It shows that the measured reduction did not require selecting a sentence owned by a C8 record; it does not show which selected sentences produced the change.

\subsection*{Brief and semantically isolated submissions receive less coverage}

Brief and semantically isolated responses had lower coverage in both topics. Records in the bottom word-count quintile had mean coverage differences of $-0.149$ in Education (95\% bootstrap CI [$-0.159$, $-0.140$]) and $-0.143$ in Trust [$-0.154$, $-0.133$]. Records in the top two semantic-isolation quintiles had differences of $-0.152$ [$-0.159$, $-0.144$] and $-0.134$ [$-0.142$, $-0.127$]. Records classified by the prespecified lexical rules as assertive had higher observed mean coverage than all other records ($+0.104$ in Education and $+0.097$ in Trust). Records classified as hedged had slightly lower observed mean coverage in Education ($-0.023$ [$-0.040$, $-0.006$]), while the corresponding contrast in Trust was inconclusive ($-0.014$ [$-0.030$, $0.003$]; Table S4). These comparisons describe observed writing features. The groups are defined from the texts themselves, so the differences are not causal effects.

\begin{figure}[ht!]
  \centering
  \includegraphics[width=\linewidth]{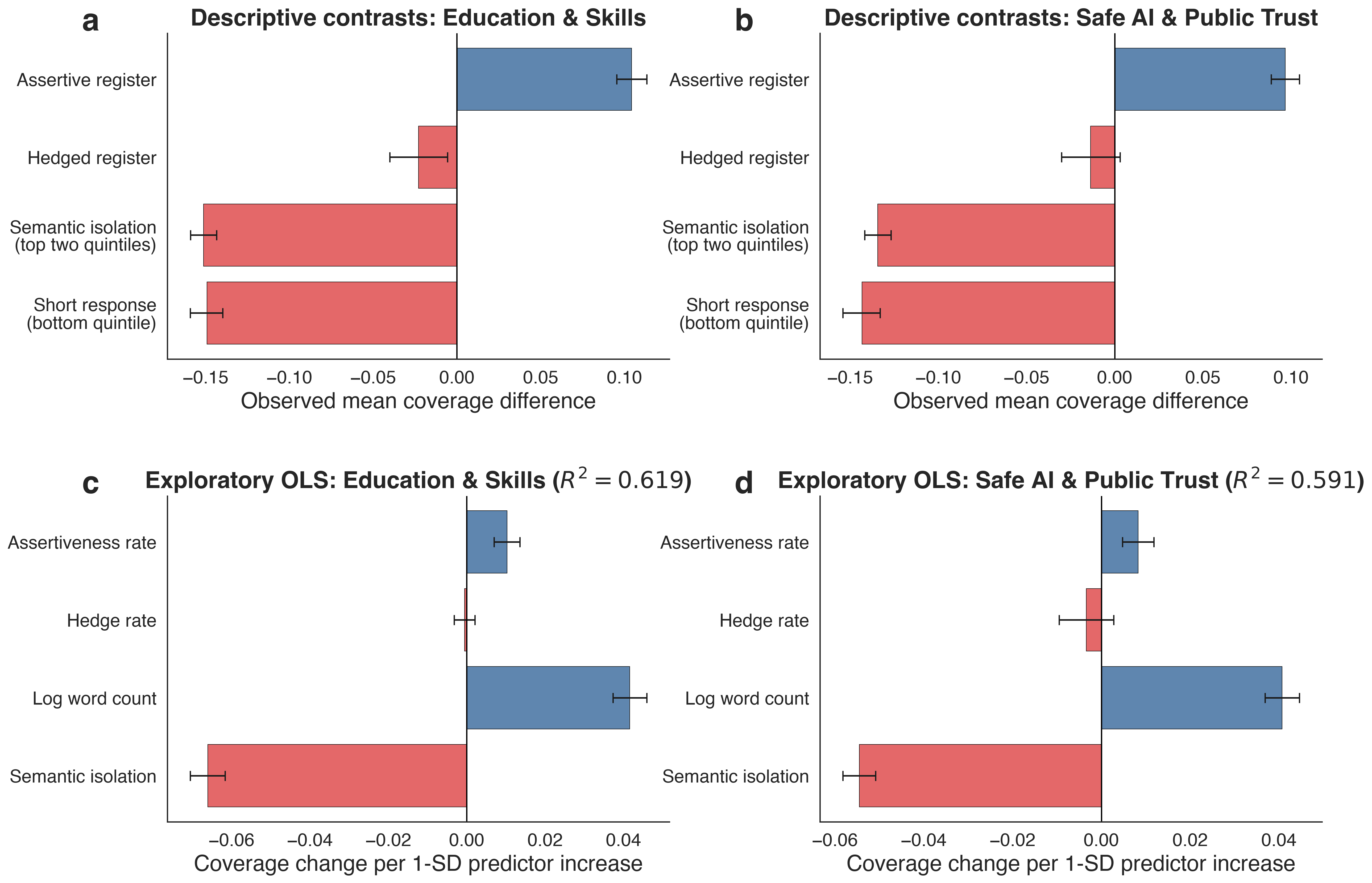}
  \caption{\textbf{Observed text features are associated with semantic coverage.} \textbf{a,b,} Unadjusted group-minus-complement differences in mean coverage with 95\% stratified-bootstrap intervals. Each contrast compares the named group with all remaining records. \textbf{c,d,} Coefficients from exploratory multivariable OLS with 95\% HC3 heteroskedasticity-robust intervals. Predictors, but not the coverage outcome, were standardized; each coefficient therefore represents the coverage-score difference associated with a one-standard-deviation increase in the predictor. These estimates are descriptive and associational, not causal effects. Semantic isolation and coverage were calculated from the same embedding vectors, so part of their association follows from the measure itself.}
  \label{fig:associations}
\end{figure}

Exploratory OLS with standardized continuous predictors and HC3 standard errors showed the same directions (Fig.~\ref{fig:associations}; Table S5). A one-standard-deviation increase in log word count was associated with $+0.042$ coverage in Education and $+0.041$ in Trust. The coefficients for semantic isolation were $-0.066$ and $-0.055$, and those for assertiveness rate were $+0.010$ and $+0.008$. Hedge rate was not distinguishable from zero. The directions were unchanged in models with cluster fixed effects. Semantic isolation and coverage are calculated from the same embeddings, so part of their association follows from the geometry of the measure. It should not be interpreted as a behavioral mechanism. Figure S4 reports descriptive cell means for response length and isolation without an inferential or causal interaction claim.

\subsection*{Cross-topic and embedding-model checks}
\label{subsec:robustness}

\begin{figure}[ht!]
  \centering
  \includegraphics[width=\linewidth]{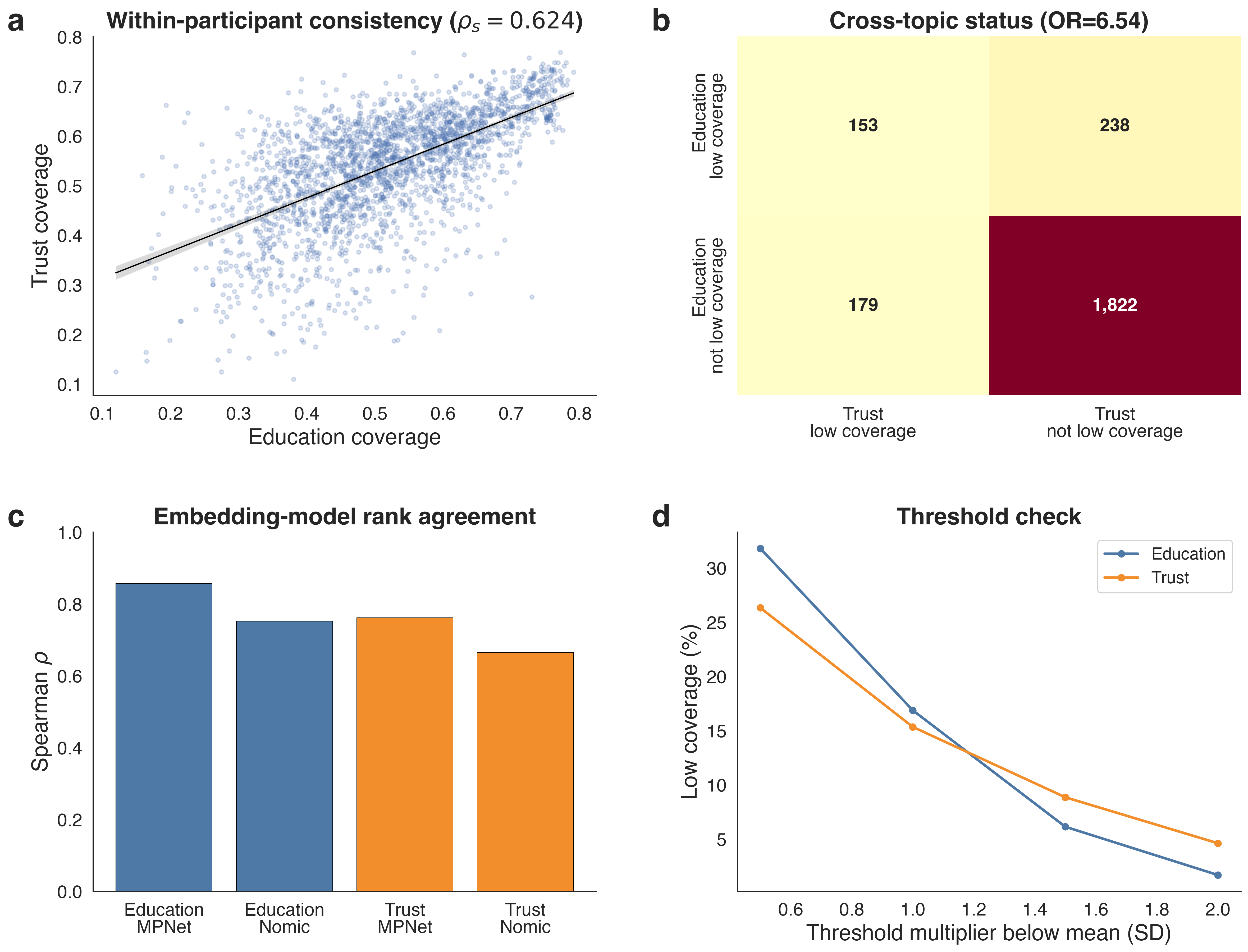}
  \caption{\textbf{Cross-topic and specification checks.} \textbf{a,} Coverage among respondents present in both topics; $\rho_s$ denotes Spearman's rank correlation. \textbf{b,} Cross-topic low-coverage contingency; OR denotes the odds ratio. \textbf{c,} Rank correlation between coverage from text-embedding-3-large and two alternative embedding models. \textbf{d,} Overall low-coverage rate under thresholds from $\bar c-0.5\sigma_c$ to $\bar c-2\sigma_c$.}
  \label{fig:robustness}
\end{figure}

The pattern was not confined to one topic. Among the 2{,}392 respondents represented in both topics, coverage scores correlated at Spearman $\rho_s=0.624$ ($p<0.001$; Fig.~\ref{fig:robustness}a). A total of 153 participants were below the threshold in both topics, 238 only in Education, 179 only in Trust, and 1{,}822 in neither. Trust low coverage occurred for 39.1\% of respondents with Education low coverage and 8.9\% of those without it. The corresponding odds ratio was 6.54 ($\phi=0.323$, $\chi^2=249.3$; Fig.~\ref{fig:robustness}b; Table S10). The association does not identify a language mechanism or explain why the same respondents had low coverage, but it shows that their relative position was not unique to one topic in this consultation.

The pattern was also not limited to text-embedding-3-large. Coverage rankings from that model correlated with all-mpnet-base-v2 at 0.857 in Education and 0.761 in Trust, and with nomic-embed-text at 0.752 and 0.665. Agreement on low-coverage status ranged from 86.3\% to 91.0\% (Fig.~\ref{fig:robustness}c; Table S11). These results indicate broad, but not identical, agreement across the three embedding models. Changing the threshold altered the overall low-coverage rate, as expected, but the same clusters remained the most affected (Fig.~\ref{fig:robustness}d; Figure S3).

\section*{DISCUSSION}
\label{sec:discussion}

The main finding is that higher observed mean coverage than a chance benchmark can coexist with large differences in whose views sit closest to a short official summary. Both official summaries had higher observed mean coverage than the exact-length random benchmark under both tested models, although the chance comparison was statistically significant only with text-embedding-3-large; this benchmark is a chance test built from contiguous word windows, not a comparison with readable alternatives. At the same time, cross-fitted benchmarks achieved better mean and lower-tail coverage without receiving more words, and the coverage left unrealized by the official summaries was concentrated rather than evenly spread. Regions centered on criticism of educational technology and distrust of technology and oversight had the largest shortfalls.

Participatory provenance asks a different question from factuality. Factuality asks whether the statements in a summary are supported by the source material. Participatory provenance asks how the summary's coverage is distributed across that material. A summary can accurately describe common themes while remaining distant from many submissions in a less visible part of the corpus. The two forms of evaluation should therefore be used together. Provenance auditing does not replace factual review, qualitative policy analysis, or editorial judgment.

This distinction also defines the role of human research. The present study is a retrospective audit of published texts. It does not measure whether participants felt represented, whether readers considered the summaries fair, or whether the summaries changed trust, legitimacy, or behavior. Those questions require research with human participants because their outcomes are human judgments or responses. In this study, representation means the defined distribution of semantic coverage. It does not refer to a participant's personal experience, and it does not establish institutional intent.

The same-budget benchmark shows how the audit could inform practice. Before publishing a summary, an analyst could inspect the lowest-covered submissions, identify where low coverage is concentrated, and revise the text while keeping the original maximum sentence-level word limits. The extractive benchmarks used here are not ready-made policy summaries because they were not evaluated for coherence, redundancy, policy usefulness, or human preference. They show only that higher semantic coverage was achievable under the tested objective within the available length. A real workflow should retain human editorial responsibility and use coverage as one check among several. If six sentences cannot cover the full range of views, a separate section for disagreement, an appendix, links from summary claims to source submissions, or a longer report may be appropriate.

The cluster results require careful interpretation. Low coverage did not map cleanly onto a simple supportive-versus-critical distinction. The analysis does not show intentional suppression, nor does it establish that every low-similarity response was absent in substance. It shows that some parts of the source material, including several regions centered on criticism and distrust, had consistently lower measured proximity to the published summaries across models and settings. Notably, one official Trust summary sentence (S6; Table S12) acknowledges skepticism toward generative AI. The 88.1\% low-coverage rate in C4 nevertheless persisted. Within the semantic-coverage measure, this pattern is consistent with acknowledging a broad theme without closely matching many submissions in that region.

The same respondents were more likely to have low coverage in both topics, which raises a wider question about inequality in participation~\citep{verba1995voice,soss2002unwanted}. Brief and semantically unusual submissions may be especially easy to lose when many responses are compressed into a small amount of text. The available data did not contain a complete set of demographic attributes, so the analysis cannot determine whether these textual patterns correspond to social disadvantage. Future studies could combine provenance measures with consented demographic data, multilingual analysis, human-coded judgments, and prospective use across institutions. Human coding would provide a complementary measure of representation rather than a single ground truth. Such studies should report their coding protocol and an appropriate agreement statistic~\citep{fleiss1971measuring,artstein2008inter}.

The question is also timely. The audited consultation itself used an AI-assisted classification and summarization pipeline with human review~\citep{canada2025aistrategy}, within a broader expansion of institutionalized citizen participation~\citep{oecd2020innovative}. The EU AI Act includes transparency obligations for certain AI-generated outputs and accuracy requirements for high-risk systems~\citep{eu2024aiact}, while Canada's proposed but unenacted Artificial Intelligence and Data Act focused on system-level accountability~\citep{canada2022aida}. These instruments do not specify participant-level reporting of how evenly a synthesis covers its inputs. Publishing coverage diagnostics alongside consultation reports would make that dimension visible and auditable.

Several limitations bound the interpretation. The analysis covers two topics from one national consultation and cannot attribute a pattern to a particular model, prompt, human editor, or institutional decision. Semantic similarity is a proxy for representation: a faithful abstraction phrased in different language can receive a low score, and embedding models carry their own assumptions about similarity. Agreement across three models reduces dependence on any single model but does not remove it. Language and relevance filters may also remove valid submissions. The analysis is limited to the retained English-language records; the filters removed a small share of records and retained uncertain cases by default, but no independently human-coded sample was available to estimate filtering error.

Statistical and design choices bound the inference in specific ways. Because the $\bar c-\sigma_c$ threshold is relative, the interpretation gives greater weight to cluster concentration, continuous scores, bottom-decile means, and threshold sweeps than to the overall low-coverage rate. The optimized benchmark selects complete sentences from training participants only and improves coverage without optimizing coherence. It is a test of feasible semantic coverage under stated constraints, not a candidate replacement for the official text, which could draw on the full corpus. The inferential results rest on five cross-fitting repetitions, conditional participant-level bootstrap intervals, paired randomization tests, and repeated-partition sensitivity summaries. The bootstrap treats participant records as the sampling unit and conditions on the fitted selections and generated embeddings, so it does not include uncertainty from preprocessing, embedding generation, or refitting the selection procedure. Full-corpus clusters serve only as post hoc regional diagnostics in the benchmark analysis, and evidence from additional independent consultations is needed before the concentration pattern can be treated as general.

These limitations identify clear directions for further research, including human-coded measures of representation, consented demographic linkage, and prospective use during summary drafting. The released framework provides a reproducible template that can be adapted to consultations with response-level text and a published summary. The present study establishes a narrower point: under an explicit semantic-coverage definition, the distribution of coverage in a short official synthesis can be evaluated rather than assumed. In this consultation, the largest measured shortfalls occurred in regions centered on criticism and distrust, and cross-fitted benchmark selections achieved lower measured shortfalls within the same sentence-level word limits.

\section*{METHODS}
\label{sec:methods}

\subsection*{Data and preprocessing}

Data came from Canada's 2025 public consultation on its next AI Strategy. The analysis covers Education \& Skills and Safe AI \& Public Trust. Responses to up to three sub-questions were concatenated within respondent and topic. Official summaries were transcribed from the published ISED report and split into six sentences (Table S12)~\citep{canada2025aistrategy}.

Preprocessing applied four sequential filters. A French-language heuristic removed texts in which more than 8\% of tokens matched a curated French stop-word list (71 Education; 83 Trust). Texts shorter than five whitespace-tokenized words were removed (91 Education; 91 Trust). Spam and placeholder rules removed two records per topic. A two-zone relevance screen embedded each response locally with all-mpnet-base-v2 and compared it with the three verbatim consultation questions and three inclusive topic anchors. Scores below 0.10 were rejected, scores at or above 0.20 were accepted, and scores between those values were adjudicated by GPT-4o-mini at temperature 0. The instructions required retention of critical or premise-rejecting responses, and errors defaulted to retention. This stage removed 50 Education and 43 Trust records. The thresholds were conservative round values fixed before the analysis. No independently human-coded sample was available to estimate classification error (Note S1).

\subsection*{Embeddings and topology}

Participant texts and official sentences were embedded with text-embedding-3-large (3{,}072 dimensions)~\citep{openai2024embeddings}. Vectors were $\ell_2$-normalized for cosine calculations. Replication used all-mpnet-base-v2~\citep{reimers2019sentence}, and the broader model check also used nomic-embed-text~\citep{nussbaum2024nomic}. These alternatives tested whether the findings depended on one embedding model, a known concern in embedding evaluation~\citep{muennighoff2023mteb}. No generative model was used to construct the benchmark summaries.

For topology and transport, the stacked participant and official-sentence embeddings were reduced to 50 principal components, explaining 55.8\% and 56.4\% of variance. $k$-means was implemented in scikit-learn with 10 restarts over $k=4,\ldots,15$~\citep{pedregosa2011scikit}. Selection combined silhouette and gap-statistic evidence with Calinski--Harabasz and Davies--Bouldin preferences~\citep{rousseeuw1987silhouettes,tibshirani2001estimating}, followed by 100 stability subsamples containing 80\% of records; stability was summarized with the adjusted Rand index~\citep{hubert1985comparing}. Trust used a documented stability-first override from the unstable consensus $k=9$ to $k=7$. Labels were produced by GPT-4o-mini~\citep{openai2024gpt4omini} from ten centroid-nearest documents and cross-described with KeyBERT phrases~\citep{grootendorst2020keybert}; normalized pointwise mutual information summarized lexical coherence~\citep{bouma2009normalized}. Selected silhouette, stability, and coherence diagnostics are shown in Figure S1; the remaining selection metrics are retained in the machine-readable topology outputs.

For visualization in Figure~\ref{fig:overview}, the combined participant and official-sentence PCA-50 embeddings were projected into two dimensions using Uniform Manifold Approximation and Projection (UMAP), with 30 nearest neighbors, a minimum distance of 0.3, and a fixed random seed of 42~\citep{mcinnes2018umap}. UMAP was used only for visualization. Cluster assignment, semantic isolation, coverage, and inferential analyses were not calculated in this two-dimensional projection.

\subsection*{Coverage, low-coverage criterion, inequality, and transport}

Let $i$ index respondent--topic records and let $j\in\{1,\ldots,6\}$ index the six official-summary sentences. Let $\mathbf e_i$ denote the embedding of record $i$, let $\mathbf s_j$ denote the embedding of summary sentence $j$, and let $\cos(\cdot,\cdot)$ denote cosine similarity. Semantic coverage was
\begin{equation}
c(i)=\max_{j\in\{1,\ldots,6\}}\cos(\mathbf e_i,\mathbf s_j).
\end{equation}
The operational low-coverage threshold was $\tau=\bar c-\sigma_c$, where $\bar c$ and $\sigma_c$ denote the mean and standard deviation of the official-summary coverage scores within each topic and embedding model. The threshold was calculated from the official-summary distribution and held fixed when alternative summaries were evaluated. The one-standard-deviation cutoff was used as an interpretable relative screening rule rather than as a natural or validated boundary of representation. A relative rule avoids imposing one absolute cosine-similarity threshold across embedding models whose score distributions differ. Conclusions were therefore evaluated alongside continuous coverage scores, cluster concentration, bottom-decile means, and threshold-sensitivity analyses.

Lower-tail coverage was operationalized as the mean coverage among the 10\% of respondent--topic records with the lowest coverage scores. This rank-based measure is distinct from the operational low-coverage indicator $c(i)<\tau$: the former summarizes performance for a fixed least-covered fraction, whereas the latter identifies records below a threshold defined by the official-summary coverage distribution. The raw non-negative coverage Gini was
\begin{equation}
G=\frac{\sum_i\sum_l|c_i-c_l|}{2n^2\bar c}.
\end{equation}
Here, $n$ is the number of respondent--topic records, $c_i=c(i)$ and $c_l=c(l)$ are the coverage scores for records $i$ and $l$, and $\bar c$ is their mean. A Gini coefficient of zero indicates equal coverage scores across records; larger values indicate greater inequality in the coverage distribution.
No translation or minimum shift was applied because the Gini is translation-sensitive. The global distributional gap was the Wasserstein-2 distance between uniform empirical measures over participant and official-sentence PCA-50 embeddings, solved with Python Optimal Transport~\citep{villani2009optimal,peyre2019computational,flamary2021pot}. It is a geometric transport cost, not an estimate of editorial or cognitive effort. Six-centroid and greedy six-quote distances are retained only as descriptive geometric references (Table S13), not chance tests. Mean coverage, Gini, and low-coverage-rate intervals used 2{,}000 participant bootstrap resamples. Cluster association used a one-way $F$ statistic against 10{,}000 label permutations with the $+1$ correction~\citep{phipson2010permutation}.

\subsection*{Exact-length and cross-fitted budget-matched benchmarks}

The exact-length null generated 1{,}000 pseudo-summaries per topic. For each of the six official sentence word counts, a participant record at least that long was sampled without replacement within the draw and a contiguous window of exactly that length was selected at random. Coverage was recomputed for every complete pseudo-summary with each tested embedding model. One-sided empirical $p$-values test whether random summaries attain mean or bottom-decile coverage at least as high as the official summary. These tests are reported unadjusted as chance diagnostics.

The same-budget benchmark used complete participant sentences of at least six words. For an official slot of $b_j$ whitespace-tokenized words, eligible candidates contained from $\max(6,\lfloor0.75b_j\rfloor)$ through $b_j$ words and never exceeded the slot maximum. Exact duplicate sentences appearing across records and sentences identical to consultation prompts were excluded. A summary used six distinct participant owners. Candidate selection was greedy: one method maximized training mean coverage; the tail-oriented method maximized $0.5\times$ mean coverage $+0.5\times$ bottom-decile mean coverage. Candidate prefiltering retained sentences with high training-set mean or bottom-decile similarity. Neither fold assignment nor candidate prefiltering used the full-corpus cluster partition. No coherence objective was used.

Evaluation used five repetitions of shuffled five-fold cross-fitting, following the general principle that model or summary selection must remain inside the training portion of an out-of-sample assessment~\citep{stone1974crossvalidatory,varma2006bias}. In every repetition, each participant appeared once in a held-out test fold; candidates were owned only by training participants, and selection never accessed test embeddings or outcomes. Fold assignment was independent of the full-corpus cluster labels. This design tests held-out performance within the analyzed corpora and does not establish external generalization to other consultations. Predictions were averaged across the five out-of-fold repetitions at participant level. The prespecified inferential family comprised eight directional comparisons: mean-optimized versus official mean coverage and tail-optimized versus official bottom-decile mean, for two topics and two embedding models. Conditional difference intervals used 5{,}000 participant bootstrap resamples. One-sided paired randomization tests swapped official and benchmark scores within participants under the null; their $p$-values were Holm-adjusted across the eight comparisons~\citep{holm1979simple}. Because overlapping training sets make naive fold-level variance estimates unreliable, neither folds nor repeated partitions were treated as independent replicates~\citep{bengio2004unbiased}. The observed ranges across five complete repetitions were retained as partition-sensitivity summaries. Code checks verified one test prediction per participant per repetition, deterministic reconstruction of each training-only selection, no candidate-owner/test overlap, six distinct owners, unique IDs, matching embedding-file text digests, and no missing predictions (Note S2). The estimands and limits of interpretation are detailed in Note S3.

\subsection*{Associational analysis}

Four pre-specified descriptive contrasts compared bottom-word-count-quintile records, top-two-isolation-quintile records, assertive-register records, and hedged-register records with their complements. Semantic isolation was Euclidean distance from a record to its assigned $k$-means centroid in PCA-50 space; larger values indicate that a record lies farther from the central content of its assigned semantic region. The register categories were dictionary-based descriptions of lexical patterns, not human-coded judgments of tone or intent. Assertiveness markers included directive or certainty terms such as ``must,'' ``should,'' ``need to,'' ``urgent,'' and ``clearly.'' Hedge markers included qualifying terms such as ``may,'' ``might,'' ``could,'' ``perhaps,'' and ``possibly.'' Assertiveness and hedge rates were calculated as the corresponding pattern-match counts divided by whitespace word count. A record was labelled assertive when its assertiveness rate exceeded 1.5 times its hedge rate, hedged when its hedge rate exceeded 1.5 times its assertiveness rate, and mixed otherwise. These labels refer only to the operational lexical rules. Differences in mean coverage used 2{,}000 stratified-within-group bootstrap resamples and Welch tests. These are observed group contrasts, not causal effects: the text-derived groups are not interventions, and neither exchangeability nor a defensible missing-data or treatment-assignment model is established~\citep{robins1994estimation,scharfstein1999adjusting,imbens2015causal}.

Exploratory OLS regressed coverage on standardized log word count, semantic isolation, hedge rate, and assertiveness rate with HC3 heteroskedasticity-robust standard errors~\citep{mackinnon1985heteroskedasticity}. A sensitivity model added cluster indicators. Because semantic isolation and coverage are derived from the same embedding vectors, the coefficients are associational diagnostics.

\subsection*{Cross-topic and model checks}

Participants appearing in both corpora were matched by internal ID. Spearman correlation quantified coverage consistency; a $2\times2$ table quantified low-coverage association using Pearson $\chi^2$, $\phi$, and the odds ratio. The model checks recomputed coverage, raw Gini, and the official $\tau=\bar c-\sigma_c$ criterion with all-mpnet-base-v2 and nomic-embed-text. Parameter sweeps varied the number of PCA dimensions (20, 50, and 100), the threshold multiplier (0.5, 1.0, 1.5, and 2.0), and the number of $k$-means regions ($k^*-2$, $k^*$, and $k^*+2$), where $k^*$ denotes the selected number of regions for that topic.

\subsection*{Computational reproducibility}

For each run, the pipeline constructs one analysis-ready dataset for each topic. Each dataset combines the retained participant records, the six official-summary sentences, eligible sentence candidates, and exact-length random windows with their corresponding text-embedding-3-large and all-mpnet-base-v2 embeddings. Nomic-embed-text embeddings cover participant records and official-summary sentences for the additional model check. All later stages read these same generated files. Record counts, vector order, and text digests are checked before analysis, and deterministic stages use fixed seeds. GPT-4o-mini was used only for relevance adjudication and initial cluster-label generation. It was not used to calculate coverage or construct benchmark summaries. The benchmark selected existing participant sentences from the generated candidate corpus. The repository README gives the full execution order and notes that new external-model calls can produce small numerical differences from the reported run.

\section*{RESOURCE AVAILABILITY}

\subsection*{Data and code availability}
The analysis code and the two consultation-response CSV files used in this study are publicly available in the \href{https://github.com/sachit27/AI-provenance}{AI-provenance GitHub repository}. The source consultation dataset is also publicly available from the \href{https://open.canada.ca/data/en/dataset/bc8cdd54-19cf-4f62-a3d3-fa4b7371d49a}{Government of Canada Open Government Portal} under the Open Government Licence--Canada.

\section*{Supplemental information index}
\begin{description}
  \item Document S1. Figures S1--S4, Tables S1--S14, and Notes S1--S3
\end{description}

\newpage
\bibliography{references}
\end{document}